\documentclass[letterpaper, 10 pt, conference]{ieeeconf}  
\usepackage[utf8]{inputenc}
\usepackage{amsmath, amsfonts}
\usepackage{graphicx}
\usepackage{xcolor}
\usepackage{wrapfig}
\usepackage{adjustbox}
\usepackage{xspace}
\usepackage{epstopdf}
\usepackage{xcolor}
\usepackage{float}
\usepackage{soul}
\usepackage{hyperref}
\usepackage{siunitx}
\usepackage{algorithm}
\usepackage{algcompatible}
\usepackage[capitalize]{cleveref}
\usepackage{booktabs}
\usepackage{multirow}
\usepackage[para]{threeparttable} 
\usepackage{array}
\usepackage{colortbl}
\usepackage{siunitx}
\usepackage{pifont}

\usepackage{eqexpl}
\eqexplSetDelim{=}

\usepackage{algpseudocode}
\usepackage{array}
\usepackage{tabularx}

\usepackage{wrapfig}
\usepackage[small]{caption}
\usepackage{subcaption}
\usepackage[noadjust]{cite}
\usepackage{comment}

\IEEEoverridecommandlockouts 

\newcommand{\systemname}{\textit{}\xspace}

\title{\systemname Form-Fitting, Large-Area Sensor Mounting for Obstacle Detection}

\author{Anna Soukhovei, Carson Kohlbrenner, Caleb Escobedo,\\ Alexander Gholmieh, Alexander Dickhans, and Alessandro Roncone 
\thanks{All authors are with the University of Colorado Boulder, 1111 Engineering Drive, Boulder, CO USA. This work is partially supported by NSF FW-
HTF-R grant \#2222952.
{\tt\small name.surname@colorado.edu}.}
}
\begin{document}
\maketitle

\begin{abstract}
We introduce a low-cost method for mounting sensors onto robot links for large-area sensing coverage that does not require the sensor's positions or orientations to be calibrated before use. Using computer aided design (CAD), a robot skin covering, or skin unit, can be procedurally generated to fit around a nondevelopable surface, a 3D surface that cannot be flattened into a 2D plane without distortion, of a robot. The skin unit embeds mounts for printed circuit boards of any size to keep sensors in fixed and known locations. We demonstrate our method by constructing point cloud images of obstacles within the proximity of a Franka Research 3 robot's operational environment using an array of time of flight (ToF) imagers mounted on a printed skin unit and attached to the robot arm.  

\end{abstract}
\section{Introduction}
Robots navigate their environment using path planning, and can use spatial data as an input for obstacle avoidance. Third person cameras can be occluded and fail to report critical data during close-proximity interactions with the environment prior to contact \cite{navarro2021proximity}. Sensors mounted on a robot surface can compliment vision systems by detecting obstacles in the robot's immediate vicinity. However, manual calibration is needed for a whole-body mounting solution specific to a given robot, which can be time consuming and error prone \cite{kohlbrenner2024gentact}.  

Using sensors to guide a robot requires knowing the location and orientation of the sensors relative to a robot base. Previous robot sensing methods have relied on manual calibration \cite{kim2024armor} or external electronics, such as inertial measurement units \cite{watanabe2021self}, to determine sensor pose, which is time intensive and difficult. Instead, keeping track of sensors in a digital design stage allows physical calibration processes to be effectively circumvented.

A digital design stage can also facilitate the design of sensor mounts. Recent works have utilized 3D printed sensor mounts generated using computer aided design (CAD) that form fit to a robot surface: Massari et al. casted silicon with embedded fiber Bragg gratings using 3D printed molds \cite{massari2022functional}, and Giovinazzo 3D printed robot coverings with sensors pasted on all sides  \cite{giovinazzo2025proxyskin}. Using digital tools allows designers to freely prototype designs and form fit their skin to a robot, so generalizing digital tooling methods can enable further development and collaboration in robot skin technology. 

\begin{figure}
    \centering
\includegraphics[width=\linewidth]{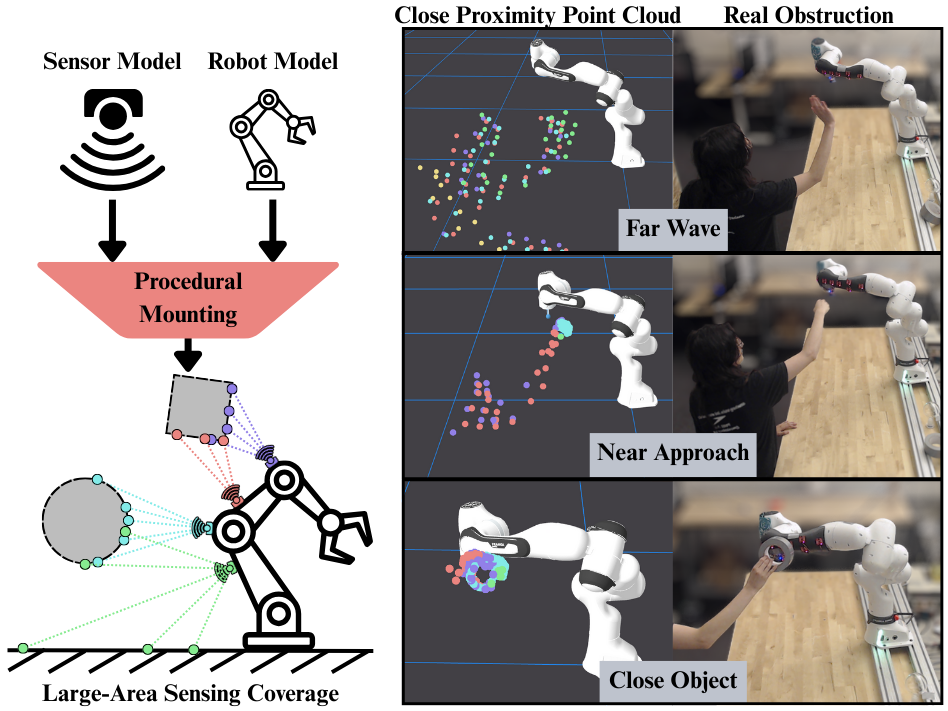}
    \caption{ (Left) A diagram showing the procedure of digitally designing a skin unit to its deployment on a robot. A model of a sensor board and a robot model are inputted into the procedural mounting step where we use 3D graphics software to fit a skin cover around a specified link. (Right) A comparison of the point cloud representation of detected obstacles versus their position relative to the FR3 in real life. Different point colors represent depth measurements from different sensors. "Far Wave" shows a person waving their hand approximately 14 cm away from the skin unit. "Near Approach" shows a person putting their fist close to the skin unit. "Close object" shows a roll of tape situated from the sensors at close proximity. In the point cloud reconstruction of the tape roll, the hollow center of the tape is clearly defined. 
}
    \label{fig:cover}
\end{figure}

 In this work, we introduce a method of procedurally generating 3D printed robot skin units that integrate sensors and provide a unique fit onto any given robot (see \cref{fig:cover}, left). Additionally, we introduce a snap-in mount for individual printed circuit boards (PCB) of sensors to attach at fixed locations on the skin unit without adhesives. The sensor mount can vary in size to accommodate different types of PCBs. We demonstrate our procedural generation algorithm and sensor mounting mechanism by manufacturing a custom skin unit for the fifth link of the Franka Research 3 (FR3) manipulator and demonstrating custom sensor mounts using time of flight (ToF) sensors. Our method builds off of Gentact Toolbox, a computational method for designing capacitative sensing skin units \cite{kohlbrenner2024gentact}. 

\section{Methods}
\label{sec:methods}
\subsection{Procedural Skin Generation}
The skin units are generated in Blender, an open-source CAD software supporting procedural generation, to generalize the design process to any robot shape. A robot model and set of vertices that represent the surface area coverage are passed into a series of operations that algorithmically output the full mount in a printable form. A thin structural layer is generated by extruding faces from the original model outwards along their local normal direction. Then, center points of the individual sensor mounts are distributed evenly on the structural layer using a Poisson disk sampling algorithm with a user defined minimum separation distance \cite{kohlbrenner2024gentact}. Requiring a minimum distance between center points maximizes sensor coverage. Finally, the 3D model of a user specified sensor's PCB is boolean subtracted at each center point, creating individual PCB mounts that allow the sensors to be snapped into place onto the skin after fabrication.
\section{Results}
\begin{figure}
    \centering
    \includegraphics[width=\linewidth]{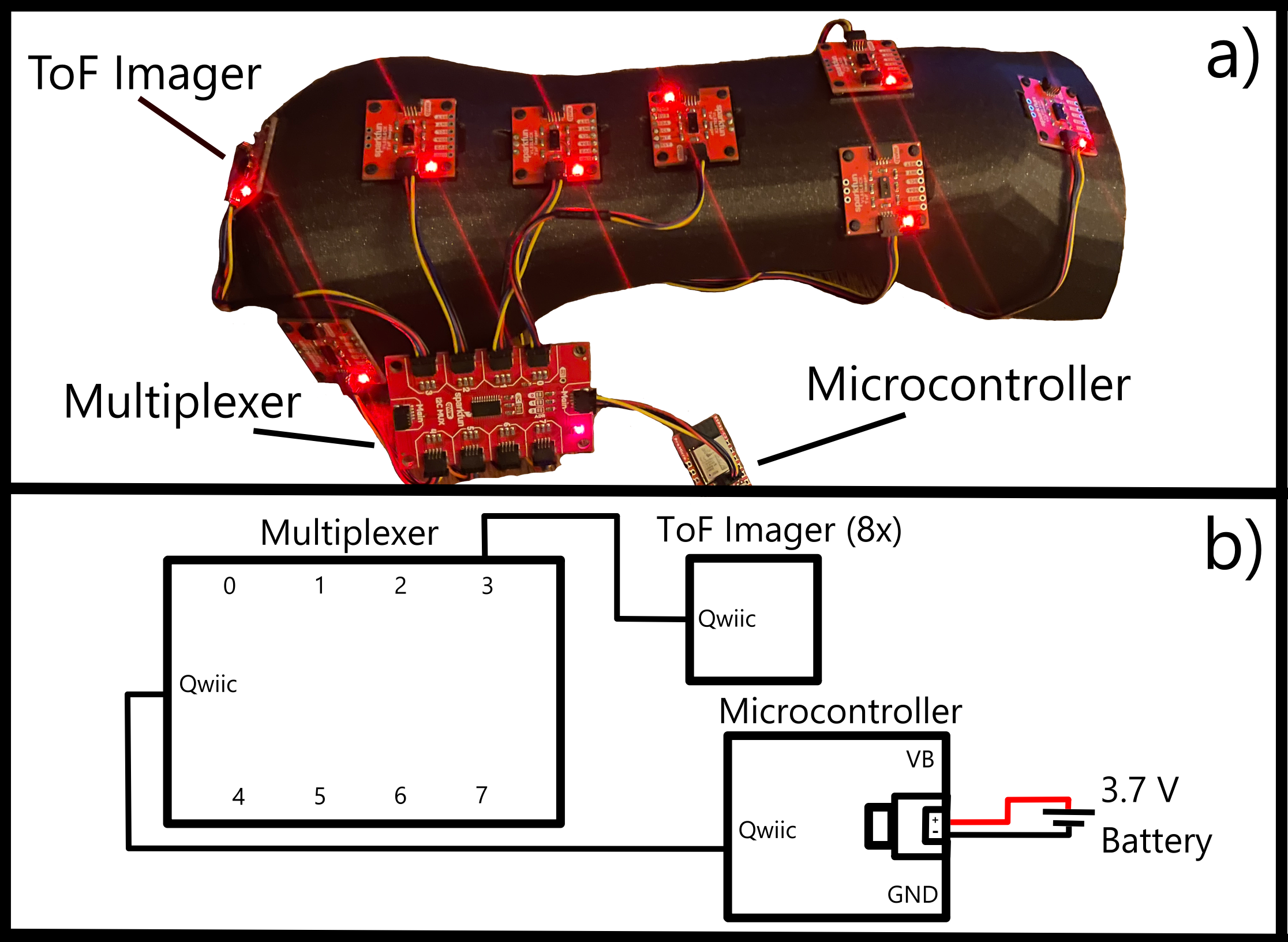}
\caption{ (a) Photo of the skin unit. For proximity sensing, we chose to mount time of flight (ToF) imagers. Each ToF imager is attached to an index (0-7) on the multiplexer. (b) General schematic of the electrical hardware.}    
    \label{fig:design-chart}
\end{figure}

\subsection{Hardware}
The skin units are fabricated using fused deposition modeling (FDM) 3D printing to generalize the process for any skin unit regardless of robot shape. Sparkfun VL53L5CX time of flight (ToF) Imagers were chosen due to their light weight specifications for prototyping: 64 zone distance measures and range accuracy of $\pm$ 15 mm for distances up to 200 mm \cite{VL53L5CX}. We snapped the ToF imagers onto the skin unit and then connected each sensor to a SparkFun Qwiic breakout multiplexer (MUX) with eight channels to aggregate data from all sensors $I^2C$ buses (see \cref{fig:design-chart}, b). The MUX main Qwiic connector is connected to both a Sparkfun ESP32-C6 microcontroller and a step-down converter to support the power draw of eight ToF imagers simultaneously. The distance data is then transmitted from the microcontroller to a ROS publisher node over WiFi using an uncompressed UDP protocol. 

To obtain a point-cloud reconstruction of surrounding objects, we transformed the distances from the ToF imagers into local frames relative to their mounting positions. For each imager, an 8$\times$8 matrix of distances in meters are processed into a viewing frustrum with a rectangular diagonal FoV of 65 degrees to project distance data along the correct outward direction.  

\subsection{Printed Skin Unit}
A Prusa XL 3D printer with 145.18 g of polylactic acid (PLA) filament was used to create a skin unit that fits on the fifth link of the FR3 (see \cref{fig:design-chart}, a). The total cost of filament was \$3.70.
The skin's dimensions are: 28 cm lengthwise, 3.81 cm tall, 10.4 cm wide, and 5 mm thick. Specifying the minimum distance between sensors as 4.5 cm, this skin is adorned with eight ToF imagers to get a wide range of overlapping coverage between sensors.

The skin unit is able to capture depth information on objects up to 3.5 meters away. In the point cloud reconstructions of obstacles, clear shapes of objects are visible: the general shape of a person, a person's arm and fist, and the 3D shape of a tape roll (see \cref{fig:cover}, right). These visualizations were obtained after more than ten removals and reapplications without manually calibrating the sensor poses. 
 \section{Conclusion}
In this work, we present a design process for large-area sensing skins that are low-cost, modular, and do not require sensor pose calibration before use. Designing the skin unit to conform to the shape of a robot link in the digital design step ensures a unique fit that secures the skin onto the robot. Furthermore, attaching sensors onto a tight fitting mount reduces the positional uncertainty of each sensor on the skin. We demonstrated that obstacles in a robot's operating environment could be reconstructed as a physically-accurate point cloud using our method. Future steps will involve quantifying the accuracy of point clouds detecting objects in close proximity to a robot base. Additionally, we plan to print skin units for all six links of the FR3 and other robots to further demonstrate that our method of procedurally generating skins is generalizable to any robot form. We hope this research can help lower the barrier of entry to whole body coverage of a robot in various modes of sensing.

\bibliographystyle{IEEEtran}
\bibliography{references}
\end{document}